\documentclass[conference]{IEEEtran}
\usepackage{times}

\usepackage[numbers]{natbib}
\usepackage{multicol}
\usepackage{balance}
\usepackage[bookmarks=true]{hyperref}
\usepackage{amsthm}

\usepackage{jparticle}
\usepackage{diagbox}
\usepackage{stfloats}

\newcommand{\GAPS}{\texttt{M-GAPS}}
\newcommand{\DiffTune}{\texttt{DiffTune}}
\newcommand{\OnePoint}{\texttt{OPRF}}

\DeclareMathOperator{\softclamp}{softclamp}

\newcommand{\kpxy}{k_p^{\mathrm{xy}}}
\newcommand{\kixy}{k_i^{\mathrm{xy}}}

\newcommand{\kvxy}{k_v^{\mathrm{xy}}}
\newcommand{\krxy}{k_r^{\mathrm{xy}}}
\newcommand{\kwxy}{k_\omega^{\mathrm{xy}}}
\newcommand{\kpz}{k_p^{\mathrm{z}}}
\newcommand{\kiz}{k_i^{\mathrm{z}}}

\newcommand{\kvz}{k_v^{\mathrm{z}}}
\newcommand{\krz}{k_r^{\mathrm{z}}}
\newcommand{\kwz}{k_\omega^{\mathrm{z}}}

\renewcommand{\th}{\xi}     %
\newcommand{\tq}{\tau}    %
\newcommand{\logR}{r}     %
\newcommand{\dt}{\delta}  %

\newcommand{\Gpg}{G_{\mathrm{pg}}}

\newcommand{\ierr}{\overline i}  %

\newcommand{\thetamanual}{\theta^m}

\newcommand{\tdes}{z}

\newcommand{\figbadinitfigeight}{
\begin{figure*}
    \centering
    \includegraphics[width=\textwidth]{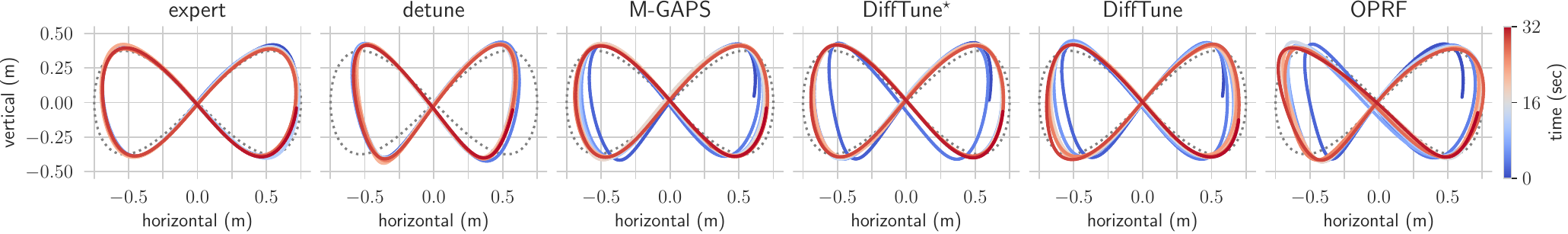}
    \caption{
        Trajectories of  a quadrotor tracking an aggressive \mbox{figure-8} trajectory under online policy optimization algorithms and fixed-parameter baselines.
        All algorithms initialized with detuned parameter.
        Dotted line shows target.
        Color changes from blue (beginning) to red (end) over time.
        See \Cref{tab:optimizers} for descriptions of optimizer names.
    }
    \label{fig:bad-init-fig8}
\end{figure*}
}

\newcommand{\figbadinitcost}{
\begin{figure}
    \centering
    \includegraphics[width=1.0\columnwidth]{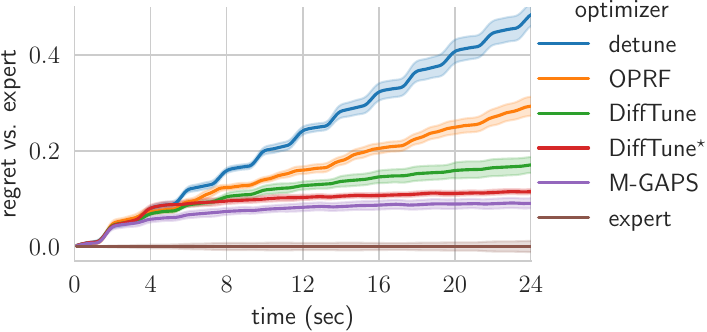}
    \caption{
        ``Regret'' (cumulative cost difference) versus expert-tuned parameters $\thetamanual$
        in figure-8 tracking experiment with detuned initialization.
        Error bars indicate $\pm 1$ standard deviation over 5 trials.
        See \Cref{tab:optimizers} for descriptions of optimizer names.
    }
    \label{fig:bad-init-cost}
\end{figure}
}

\newcommand{\figbadinitparams}{
\begin{figure}
    \centering
    \includegraphics[width=\columnwidth]{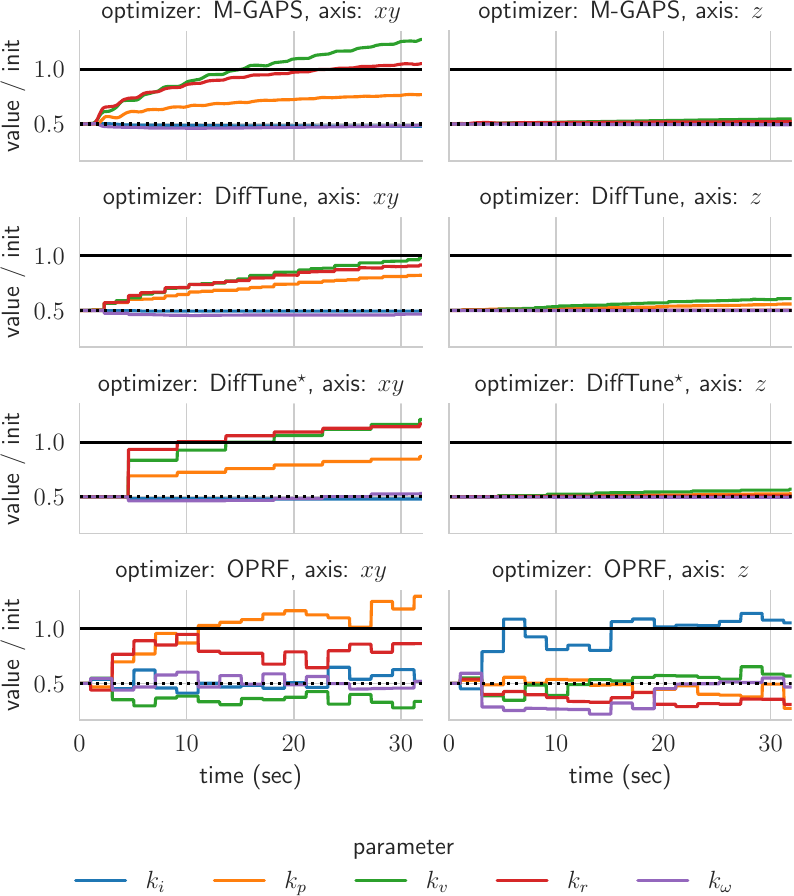}
    \caption{
        Parameter evolution in figure-8 tracking experiment with detuned initialization.
        Ratio of each parameter entry to the expert-tuned baseline $\exp(\thetamanual)$ is shown to account for widely-varying parameter magnitudes.
        \emph{Row}: Optimizer.
        \emph{Left}: Gains for horizontal motion axes.
        \emph{Right}: Gains for vertical motion axis.
        See \Cref{tab:optimizers} for descriptions of optimizer names.
    }
    \label{fig:bad-init-params}
\end{figure}
}

\newcommand{\figepisodic}{
\begin{figure}
    \centering
    \includegraphics[height=1.5in]{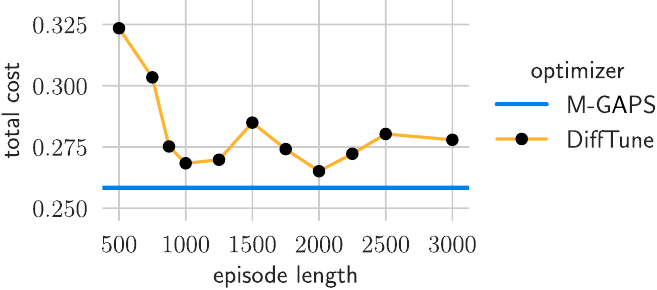}
    \caption{
        Total cost of \GAPS\ vs. \DiffTune\ with various episode lengths in detuned initialization experiment.
        \DiffTune\ with optimal episode length performs nearly as well as \GAPS, but degrades with other episode lengths.}
    \label{fig:episodic}
\end{figure}
}

\newcommand{\figweightcost}{
\begin{figure}
    \centering
    \includegraphics[width=0.85\columnwidth]{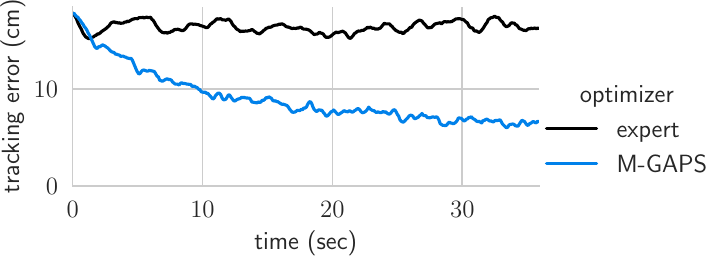}
    \caption{
        Position tracking error under \GAPS\ for heavy payload disturbance (\S \ref{sec:weight}).
        \GAPS\ substantially outperforms the expert $\thetamanual$ tuned for smaller disturbances.
    }
    \label{fig:weight}
\end{figure}
}

\newcommand{\figfancost}{
\begin{figure}
    \centering
    \includegraphics[width=\columnwidth]{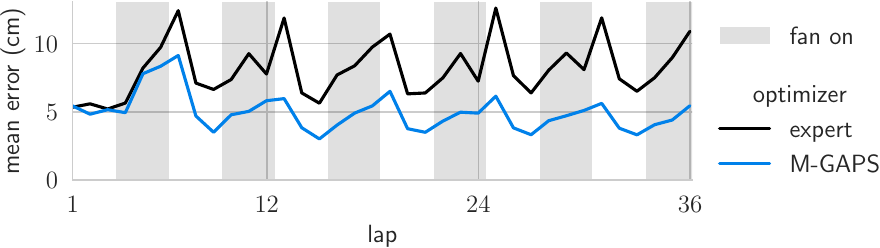}
    \caption{
        Position tracking error under \GAPS\ for periodic fan disturbance (\S \ref{sec:fan}).
        Error is averaged per ``lap'' due to within-lap variance from the airflow pattern.
        \GAPS\ substantially outperforms the expert $\thetamanual$ tuned for smaller disturbances.
    }
    \label{fig:fan-error}
\end{figure}
}

\newcommand{\figfanparams}{
\begin{figure}
    \centering
    \includegraphics[width=0.96\columnwidth]{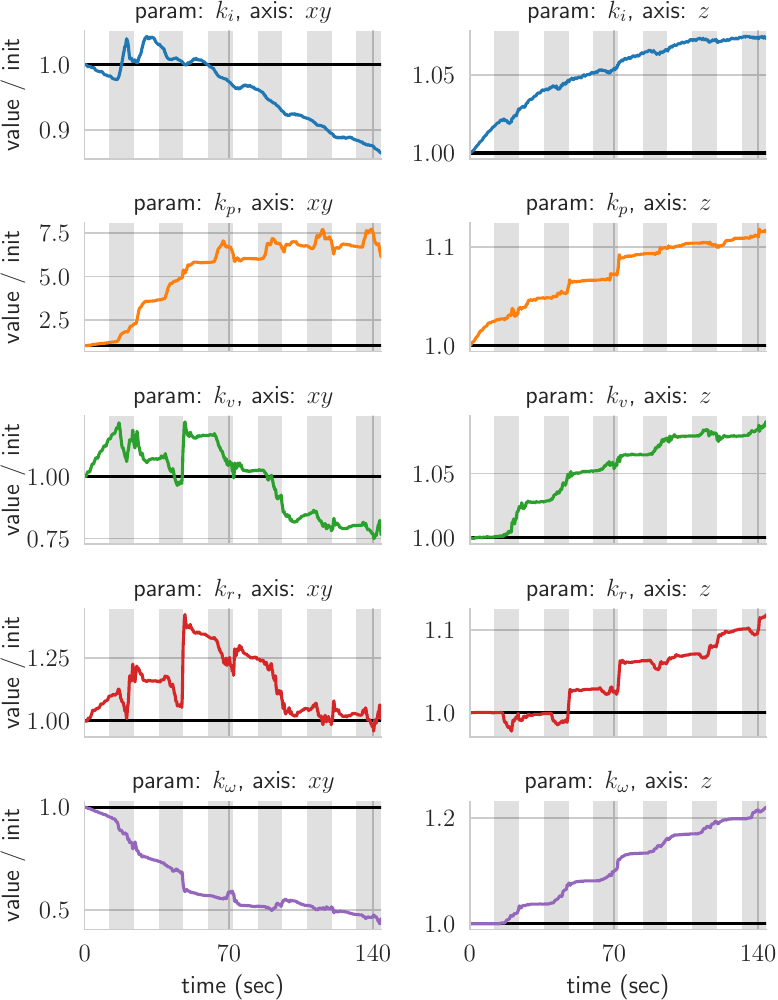}
    \caption{
        Parameter evolution under \GAPS\ for periodic fan disturbance (\S \ref{sec:fan}).
        Shaded bands indicate when the fans are energized.
        Some parameters react strongly to fan phase, showing rapid adaptation.
    }
    \label{fig:fan-params}
\end{figure}
}

\newcommand{\figcompareparams}{
\begin{figure}[b]
    \centering
    \includegraphics[width=0.95\columnwidth]{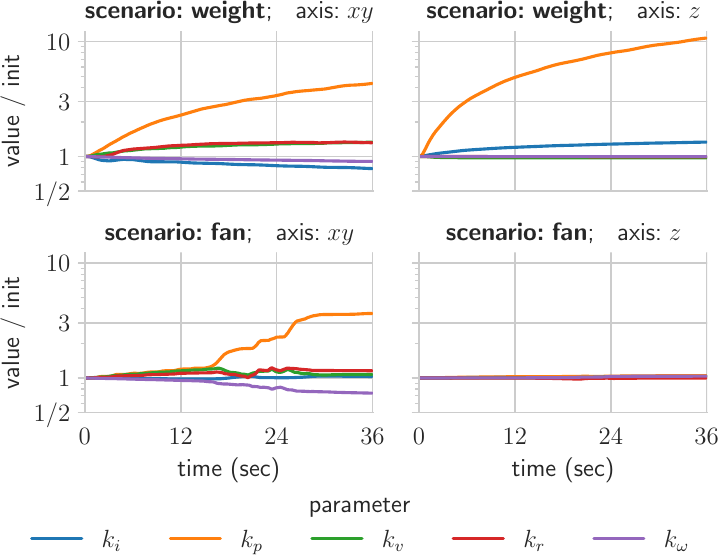}
    \caption{
        Parameter evolution under \GAPS\
        in heavy payload (top, \S \ref{sec:weight})
        and abbreviated fan (bottom, \S \ref{sec:fan}) scenarios.
        Logarithmic $y$-axis shows
        $\frac{\text{evolved parameter}}{\text{initial value}}$.
        Differences between scenarios are large,
        showing instance adaptivity of \GAPS.
    }
    \label{fig:compare-params}
\end{figure}
}

\newcommand{\figtraxxas}{
\begin{figure}
    \centering
    \includegraphics[width=0.8\columnwidth]{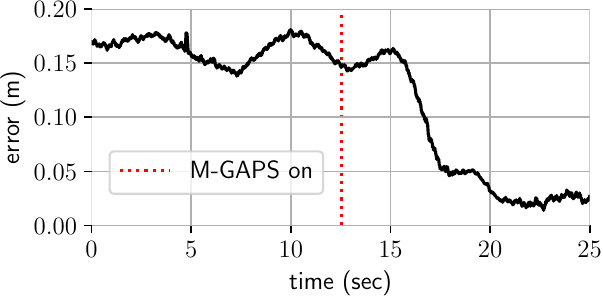}
    \caption{
        \GAPS\ tuning parameters of a nonlinear controller for an Ackermann-steered 1:6-scale car traveling in a circle with period of \SI{12.5}{sec} and radius of \SI{2.2}{m}.
        Tracking error reduced by over $5 \times$ within one period.
    }
    \label{fig:traxxas}
\end{figure}
}

\title{\LARGE \bf
Fast Non-Episodic Adaptive Tuning of Robot Controllers \\ with Online Policy Optimization
}

\author{
   \authorblockN{James A. Preiss}
   \authorblockA{
      Computer Science Department\\
      University of California, Santa Barbara\\
      Santa Barbara, California, USA\\
      preiss@ucsb.edu (corresponding author)
   }
   \and
   \authorblockN{Fengze Xie, Yiheng Lin, Adam Wierman, and Yisong Yue}
   \authorblockA{
      Department of Computing + Mathematical Sciences\\
      California Institute of Technology\\
      Pasadena, California, USA\\
      \{fxxie, yihengl, adamw, yyue\}@caltech.edu
   }
}

\begin{document}

\maketitle

\begin{abstract}
    We study online algorithms to tune the parameters of a robot controller 
    in a setting where the dynamics, policy class, and optimality objective are all time-varying.
    The system follows a single trajectory without episodes or state resets, and the time-varying information is not known in advance.
    Focusing on nonlinear geometric quadrotor controllers as a test case,
    we propose a practical implementation of a single-trajectory model-based online policy optimization algorithm, \GAPS,
	along with reparameterizations of the quadrotor state space and policy class to improve the optimization landscape.
    In hardware experiments,
	we compare to model-based and model-free baselines that impose artificial episodes.
    We show that \GAPS\ finds near-optimal parameters more quickly, especially when the episode length is not favorable.
    We also show that \GAPS\ rapidly adapts to heavy unmodeled wind and payload disturbances,
    and achieves similar strong improvement on a 1:6-scale Ackermann-steered car.
    Our results demonstrate the hardware practicality of this emerging class of online policy optimization that offers significantly more flexibility than classic adaptive control, while being more stable and data-efficient than model-free reinforcement learning.
\end{abstract}

\IEEEpeerreviewmaketitle

\section{Introduction}
\label{sec:introduction}

We study the problem of optimizing a parameterized non-linear robot control policy in an online setting.
A deployed robot may face unpredictable changes in both environment and task, and must adapt to them immediately.
Therefore, we consider a protocol where the dynamics, policy class, and cost functions are all time-varying and revealed online:
the optimization algorithm has no knowledge of how they will vary in the future.
The algorithm is evaluated on a single trajectory without episodes or state resets.
As a case study, we focus on nonlinear trajectory tracking control for quadrotors.

Policy optimization has been widely studied in the control and machine learning communities from varying perspectives.
We are interested in methods that:
\begin{enumerate}
    \item \label{des:assum}     Can be applied to general nonlinear dynamics and costs.
    \item \label{des:direct}    Optimize a given policy class (vs. prescribe their own).
    \item \label{des:adaptive}  Are adaptive -- do not rely on stationarity assumptions on the dynamics or costs.
    \item \label{des:efficient} Are efficient enough to run onboard a microcontroller.
\end{enumerate}
The criteria \ref{des:assum}-\ref{des:direct} rule out most adaptive control methods,
which involve a co-designed policy and adaptive law and have limited ability to optimize costs besides quadratic tracking \cite{slotine1991applied}.
Reinforcement learning (RL) is more general,
but the criteria \ref{des:adaptive}-\ref{des:efficient} rule out
methods but that rely on a replay memory~\cite{haarnoja-sac},
which assumes stationarity and is computationally expensive.
Policy gradient RL avoids these issues,
but popular methods assume that dynamics and costs are unknown \cite{schulman-PPO}, leading to slow, data-inefficient learning that that impedes adaptivity.

For many practical robotic systems,
adaptive control is too restrictive,
but RL is too general.
Engineers often
design the cost function
and know an approximate dynamics model,
but model-free policy gradient RL fails to use this information.
New automatic differentiation algorithms and software
have broadened the scope of differentiable 
policy classes~\citep{amos2018DiffMPC} and dynamics models~\citep{coros2021differentiable},
including
general rigid bodies \citep{deavila2018diffphysics},
finite element models \citep{xue2023JAXFEM},
and fluids \citep{schenck2018differentiable},
among many others.
However, differentiable models are still approximate, and not always useful~\citep{suh2022differentiable}.
Therefore, in this work we include an ``apples-to-apples'' comparison of model-based and model-free methods in real time on hardware.

Our protocol is structurally similar to online optimization.
In \emph{stateless} online optimization with certain convex functions, it has been shown that online gradient descent yields minimax-optimal regret \citep{hazan2022oco}.
However, in the \emph{stateful} online \emph{policy} optimization problem, it is not obvious how to apply the same algorithmic principle.
One simple approach is to artificially impose episodes, but this
fails to account for the influence of previous actions on each episode's initial state.
It also causes large  step changes in policy
and adds another hyperparameter to tune.
On the other hand, a simple non-episodic method was shown to obtain optimal regret in linear systems when the closed-loop dynamics under the policy class satisfy a form of contractiveness \citep{lin2023gaps}.
However, in nonlinear systems, the analysis only provides a \emph{local regret} bound
analogous to tracking a moving stationary point, leaving it unclear if
this method works well in practice.

\textbf{Contributions.}
We conduct real-hardware experiments comparing three algorithms that extend the online gradient descent principle to online policy optimization:
a model-free episodic method \OnePoint~\citep{zhang2024boosting},
a model-based episodic method \DiffTune~\citep{difftune},
and a model-based non-episodic method \GAPS~\citep{lin2024colt}.
We propose reparameterizations of
the quadrotor state space
and policy class
to improve the optimization landscape for all methods.
We study three scenarios: 
1) initialization with suboptimal parameters, %
2) a strong time-varying wind, and
3) a heavy unmodeled payload.
In setting~1, we find that \GAPS\ performs best overall.
\DiffTune\ is close when the episode length is optimal, but significantly worse otherwise;
\OnePoint\ lags further.
In settings~\mbox{2-3}, we confirm that \GAPS\ outperforms an expert hand-tuned baseline despite an imperfect dynamics model.
Setting~3 also demonstrates fast adaptivity to strongly time-varying dynamics.
We verify that these positive results extend to other hardware with an experiment on a 1:6-scale off-road car.
Together, our results show that the model-based, non-episodic algorithm \GAPS\ is a promising approach for general-purpose, data-efficient, and fully-online robotic controller tuning.

\paragraph*{Protocol}
We consider a discrete-time, time-varying, online process.
At timestep $t \geq 0$,
the learner observes the current state $x_t \in \cX \subseteq \R^n$.
The learner selects a policy parameter $\theta_t \in \Theta \subseteq \R^d$, which determines an action $u_t \in \R^m$ via the parameterized policy class $u_t = \pi_t(x_t, \theta_t)$.
The learner then suffers the cost $c_t = f_t(x_t, u_t) \in \R$
and the state advances according to the dynamics $x_{t+1} = g_t(x_t, u_t)$.
Finally, the learner observes the partial derivatives of $\pi_t$, $f_t$, and $g_t$ about $(x_t, u_t, \theta_t)$.
This protocol repeats along a single trajectory, without any resetting or episodes, for $T$ timesteps.
The learner's goal is to select the parameters $\theta_t$ online to minimize the total cost
$\sum_{t=0}^{T-1} c_t(x_t, u_t)$.
Note that our setting can express known time-invariant dynamics with unknown adversarial additive disturbances as a special case.

\paragraph*{Notation}
The shorthand $x_{i:j}$ refers to the sequence $x_i, \dots, x_j$.
The matrix or vector of all zeros, with dimension implied by context, is denoted by $\zero$.
The standard basis vectors of $\R^3$ are denoted by
$e_x, e_y, e_z$.
For $f : \R \mapsto \R$ and $x \in \R^n$, the overloaded $f(x)$ indicates elementwise application of $f$.

\section{Related Work}
\label{sec:related}

\paragraph{Adaptive Control}
The field of adaptive control focuses on specific structures of unknown time-varying dynamics.
The most widely studied nonlinear systems are those of the form
$x_{t+1} = g_t(x_t, u_t) + \phi(x_t)^\top a_t$,
where the nominal model $g_t$ is known,
$\phi : \R^n \mapsto \R^{n \times k}$ is a known feature mapping,
and $a_t \in \R^k$ is unknown~\cite{slotine1991applied}.
A proxy $\hat a_t$ for $a_t$ is updated by the adaptive law
and used in the policy.
In estimation-based laws, $\hat a_t$ is updated to reduce the prediction error
$\norm{x_{t+1} - g(x_t, u_t) - \phi(x_t)^\top \hat a_t}_2^2$,
which is not directly related to any optimality criterion.
However, in tracking-based laws \cite{ioannou1995robust},
$\hat a_t$ is updated to reduce the tracking error
$\norm{x_{t+1} - x^d_{t+1}}_2^2$
where $x^d$ is a desired state, %
making them interpretable as a form of policy optimization.
Composite laws combine both objectives \cite{neuralfly, krstic1995nonlinear}.

In the highly-studied ``matched-disturbance'' case, the dynamics take a restricted form that enables straightforward adaptive controllers with Lyapunov stability analysis.
Although recent work has relaxed the matched-disturbance condition \cite{lopez2021universal},
the policy class still takes a prescribed form incompatible with the quadrotor controller considered in this paper.
Also, while Lyapunov analysis delivers exponential contraction of $\norm{x_t - x^d_t}_2^2$,
it lacks guarantees for general time-varying costs.

Other tools from control theory related to our setting are
iterative learning control (ILC) and self-tuning regulators (STRs).
ILC accounts for realized (not worst-case) nonstochastic disturbances~\cite{bristow2006ILC},
but it assumes a cyclic setting,
whereas we allow arbitrary changes.
STRs perform online system identification and repeatedly solve optimal control synthesis 
on the estimated model.
This implies that control synthesis is easy given the model, which does not hold in our setting.
Many adaptive control schemes have been proposed specifically for quadrotors \citep{mo2019nonlinear},
but we consider algorithms that can be applied to general systems
and use quadrotors as a test case.

Although the methods we investigate are more expressive than typical adaptive control in terms of policy optimization, they are complementary in terms of dynamics learning.
Adaptive control with a pure estimation-based law can work \emph{in tandem} to provide online estimates of the dynamics $g_t$ consumed by the online policy optimization algorithm.
The resulting estimator-optimizer combination can be theoretically sound under assumptions not significantly stronger than those required by each in isolation~\citep{lin2024colt}.

\paragraph{Reinforcement Learning}
RL is generally divided into \emph{value-based} methods that approximate the optimal value function, and \emph{policy-based} methods that optimize a parameterized class of policies.
In the latter case, when the system model is unknown, this optimization is both stochastic and derivative-free.
Algorithms can be based on exploration in the action space (REINFORCE and descendants \citep{williams-REINFORCE,schulman-PPO}),
or in the parameter space \citep{sehnke2010parameter}.
In either case, these methods deploy the same policy parameter $\theta$ for a horizon of $H$ steps and construct an unbiased estimator of
$\nabla_\theta \sum_{s=t}^{t+H-1}f_s(x_t,u_t)$.
Typical analysis assumes the episodic structure comes from the problem itself, and the state is reset to a fixed distribution for each episode.
However, under suitable mixing assumptions \cite{baxter2001infinite} or infinite-horizon discounted objectives \cite{paternain2022continuing}, one can enact the episode divisions artificially and still obtain results in the single-trajectory setting.
But the choice of episode length becomes a hyperparameter, and these methods do not use any knowledge of the $g_t, f_t$, so they are data-hungry.

If a model is available, one can compute the episode cost gradient
$\nabla_\theta \sum_{s=t}^{t+H-1}f_s(x_t,u_t)$ analytically.
However, naive optimization in simulation without any trajectory data from the real system is prone to exploiting model errors, and may be counterproductive for stiff or discontinuous systems even when the model is correct \cite{suh2022differentiable}.
The method of \cite{difftune} uses a model-based gradient taken about the true system trajectories in an episodic manner, and is one of the baselines we study in this work.
Many other methods have been proposed to mitigate model error exploitation for policy optimization under learned models \cite{amos2021modelbased}
or to improve the optimization conditioning \cite{mora2021pods}.
Most of this work focuses on time-invariant but complex settings and is computationally expensive, whereas we consider time-varying settings and simpler algorithms.

In contrast to typical policy-based RL, value-based RL methods targeting infinite-horizon discounted objectives can often handle the single-trajectory protocol without imposing artificial episodes.
Although their objective is not formally compatible with fast-changing $g, f$, they can work in slowly time-varying settings.
Methods based on $Q$-learning struggle with real-time application in continuous action spaces because taking $\argmax_u Q(x, u)$ is a hard nonconvex optimization problem \cite{kalashnikov2018qtopt}.
A more suitable family is the actor-critic methods \cite{haarnoja-sac,silver2014deterministic}, which optimize a policy and value function simultaneously.
If a dynamics model is available, actor-critic methods can 
approximate the state-value function
instead of the more complex (state, action)-value $Q$ function.
However, even the simplest possible algorithm in this class requires many tuning and design decisions~\cite{werbos1991menu}.
For the baselines in this work, we focus on methods in the policy gradient family that are simple and computationally lightweight.

\paragraph{Online Policy Optimization}
Recently, ideas from online optimization have been applied to obtain regret-optimal algorithms for control problems with online time-varying dynamics and/or costs \cite{hazan2022nonstochasticbook}, as in our setting.
Initial work considered only adversarial disturbances and cost functions with a known linear time-invariant system~\cite{agarwal2019online}.
Although later extended to 
unknown time-invariant~\cite{hazan2020nonstochastic},
known time-varying~\cite{gradu2023adaptive},
and partially observed~\cite{simchowitz2020improper} systems,
this family prescribes specialized policy classes that depend strongly on the linearity of the dynamics.
A different line of work \cite{lin2023gaps, lin2024colt} focuses on optimizing arbitrary policy classes when the closed-loop dynamics satisfy a particular notion of contractiveness.
These methods recover optimal regret when instantiated with linear dynamics and appropriate policy classes \cite{lin2023gaps, lin2024colt},
but also provide a weaker ``local regret'' guarantee under contractive dynamics but nonconvex optimization, as elaborated in \S \ref{sec:mgaps}.
This case is especially relevant for the complex nonlinear systems encountered in robotics.
We select the algorithm of \cite{lin2024colt} for our experiments,
as it has the lightest memory and computation needs.

\section{Methods}
In this section, we define the three algorithms used in our experiments.
As discussed in \S \ref{sec:introduction}, each algorithm can be interpreted as applying the algorithmic spirit of online gradient descent to the online policy optimization problem.

\subsection{Single-trajectory method (\GAPS)}
\label{sec:mgaps}
\begin{algorithm}[t]
	\caption{Non-episodic, model-based (\GAPS\ \cite{lin2024colt}) }
	\label{alg:mgaps}
	\begin{algorithmic}
		\Require{parameter $\theta_0$, learning rate $\eta$.}
		\State $y_0 \gets \zero$.
		\For{$t \in 0, \dots, T-1$}
			\State Deploy $u_t = \pi_t(x_t, \theta_t)$ in environment.
			\State Observe $x_{t+1}$ and derivatives of $g_t$ and $f_t$.
			\State Compute $G_t$ as in \eqref{eq:mgaps-grad}.
			\State $\theta_{t+1} \gets \theta_t - \eta G_t$.
			\State Update $y_t$ as in \eqref{eq:mgaps-y}.
		\EndFor
	\end{algorithmic}
\end{algorithm}
The non-episodic online policy optimization algorithm \GAPS\ was introduced and analyzed in a meta-framework alongside online dynamics estimation~\citep{lin2024colt}.
Here, we consider a simplified case where (approximate) dynamics are known and the parameter set $\Theta$ is an unconstrained Euclidean space.
\GAPS\ provides theoretical guarantees in nonlinear, nonconvex settings.
To state these guarantees precisely, we introduce the concept of \emph{surrogate cost} \citep{lin2023gaps,lin2024colt} to evaluate a policy parameter $\theta$ at a particular timestep.
The surrogate cost $F_t : \Theta \mapsto \R$ for timestep~$t$ is defined as
\(
	F_t(\theta) = f_t(x_t^\theta, u_t^\theta),
\)
where $x_t^\theta$ is the state visited at time $t$ if the policy parameter $\theta$ is deployed at all previous timesteps, i.e. if $\theta_0, \dots, \theta_{t-1} = \theta$,
and $u_t^\theta = \pi_t(x_t^\theta, \theta)$.
Note that $x_t^\theta$ and $u_t^\theta$ are both completely determined by the choice of $\theta$.
When $F_t$ is convex in $\theta$ for all timesteps $t$,
\GAPS\ targets the metric of \emph{policy regret},
\begin{equation}
\label{eq:policy-regret}
	R_P(T) = \textstyle \sum_{t=0}^{T-1} f_t(x_t, u_t) -
	\min_{\theta \in \Theta} \sum_{t=0}^{T-1} F_t(\theta),
\end{equation}
where $x_t, u_t$ are the states and actions visited by the online learner.
Examples of convex $F_t$ include linear dynamics with disturbance-action control \cite{hazan2022nonstochasticbook} or model-predictive control with disturbance predictions and confidence coefficients~\cite{lin2023gaps}.
In the convex case, the online gradient descent (OGD) update
$\theta_{t+1} \gets \theta_t - \eta \nabla F_t(\theta_t)$
with a suitable learning rate $\eta$
is minimax-optimal for minimizing $R_P$ \citep{lin2023gaps}.

However, quadrotor and car dynamics are nonlinear, making the $F_t$ nonconvex.
In this case, \GAPS\ targets the \emph{local regret}
\begin{equation}
\label{eq:local-regret}
	R_L(T) = \textstyle \sum_{t=0}^T \norm{\nabla F_t(\theta)}_2^2,
\end{equation}
an online analog to stationary-point conditions in nonconvex optimization.
In both cases, performing OGD on the surrogate costs $F_t$ is well-motivated theoretically.
However, it requires more computation and information than is practical.
To see this, we apply one step of the chain rule to $\nabla F_t(\theta)$, obtaining
\begin{equation}
\label{eq:grad-surrogate}
	\begin{split}
		\nabla F_t(\theta)
		=
        \frac{\partial f_t}{\partial u}
        \frac{\partial \pi_t}{\partial \theta}
        +
        \left( \frac{\partial f_t}{\partial x}
            +
            \frac{\partial f_t}{\partial u}
            \frac{\partial \pi_t}{\partial x}
        \right)
		\frac{\partial x_t^\theta}{\partial \theta}.
	\end{split}
\end{equation}
All derivative dependence on past dynamics and policy functions is encapsulated in the ``sensitivity'' term
$\partial x_t^\theta / \partial \theta$,
which is computed recursively via the chain rule as
$\nofrac{\partial x_0^\theta}{\partial \theta} = \zero$~and
\begin{gather}
\label{eq:sensitivity-recursion}
    \frac{\partial x_{s+1}^\theta}{\partial \theta} =
    \left(
        \frac{\partial g_s}{\partial x_s^\theta}
        + \frac{\partial g_s}{\partial u_s^\theta} \frac{\partial \pi_s}{\partial x_s^\theta}
    \right) \frac{\partial x_s^\theta}{\partial \theta}
    + \frac{\partial g_s}{\partial u_s^\theta} \frac{\partial \pi_s}{\partial \theta}.
\end{gather}
In the hypothetical case where we deploy a constant $\theta_0, \dots, \theta_{t-1} = \theta$,
the recursive step \eqref{eq:sensitivity-recursion} computes
$\partial x_t^\theta / \partial \theta$
from a previously computed
$\partial x_{t-1}^\theta / \partial \theta$
in $O(1)$ time.
However, if the online algorithm is changing $\theta$ constantly, then 
the intermediate steps used to compute
$\partial x_{t-1}^{\theta_{t-1}} / \partial \theta_{t-1}$
are useless for computing
$\partial x_t^{\theta_t} / \partial \theta_t$.
Instead, the complete state trajectory $x_0^{\theta_t}, \dots, x_t^{\theta_t}$ must be ``re-simulated''.
This incurs $\Omega(t)$ computational cost, which quickly becomes intractable.
It also requires oracle access to
$(f_s, \pi_s)_{s=0}^t$
and
$(g_s)_{s=0}^{t-1}$
at the ``re-simulated'' states,
which may be unreasonable.

Therefore, the ideal of OGD on the surrogate costs $F_t$ is not a practical algorithm.
Instead, \GAPS\ forms a computationally efficient approximation of $\nabla F_t(\theta_t)$ with error small enough to yield optimal regret.
\GAPS\ computes the recursion as in \eqref{eq:sensitivity-recursion} \emph{without} re-simulation,
essentially ``ignoring'' the fact that $\theta_t$ is changing online.
Specifically, it maintains an internal state $y_t \in \R^{n \times d}$
that approximates the sensitivity $\partial x^{\theta_t} / \partial \theta_t$,
with the dynamics
\begin{equation}
\label{eq:mgaps-y}
    y_0 = \zero,
    \quad
    y_{t+1} = \left(
        \frac{\partial g_t}{\partial x_t}
        + \frac{\partial g_t}{\partial u_t} \frac{\partial \pi_t}{\partial x_t}
    \right) y_t
    + \frac{\partial g_t}{\partial u_t} \frac{\partial \pi_t}{\partial \theta_t}.
\end{equation}
All derivatives are taken about the actually-visited trajectory $(x_s, u_s)_{s=0}^t$, so previously computed results are never invalidated and only $O(1)$ computation is used per step.
\GAPS\ then performs an approximate online gradient descent update
$\theta_{t+1} = \theta_t - \eta G_t$,
where $\eta > 0$ is the learning rate and
\begin{equation}
\label{eq:mgaps-grad}
	G_t =
	\left( 
		\frac{\partial f_t}{\partial x_t}
		+
		\frac{\partial f_t}{\partial u_t}
		\frac{\partial \pi_t}{\partial x_t}
	\right) y_t
	+ \frac{\partial f_t}{\partial u_t} \frac{\partial \pi_t}{\partial \theta_t}.
\end{equation}
The algorithm steps are summarized in \Cref{alg:mgaps}.

\subsubsection{Contractiveness and guarantees}
\GAPS's gradient approximation $G_t$ is valid when
the closed-loop dynamics of the system and policy class are \emph{contractive}.
Let $\Phi^\theta_{t|s}(x_s)$ denote the multi-step closed loop dynamics under a policy parameter~$\theta$, defined recursively as $\Phi^\theta_{s|s}(x_s) = x_s$ and
\[
	\Phi^\theta_{t|s}(x_s) =
	g_s \big(
			\Phi^\theta_{t-1|s}(x_s),
			\pi_s (\Phi^\theta_{t-1|s}(x_s), \theta )
			\big)
	\quad \forall t > s.
\]
We say the closed-loop dynamics are \emph{contractive} if there exists
$C \geq 0$ and $0 \leq \rho < 1$ such that
for all time indices $s \leq t$, parameters $\theta \in \Theta$, and state pairs $x, x'$, we have
$
\norm{
		\Phi^\theta_{t|s}(x)
		-
		\Phi^\theta_{t|s}(x')
	}
	\leq
	C \rho^{t - s} \norm{x - x'}
$.
Intuitively, each policy parameter $\theta$ steers the state towards its ``preferred trajectory'' $x_t^\theta$ with exponential convergence if deployed for several steps in a row.
If contractiveness holds,
along with
a careful choice of learning rate~$\eta$
and mild smoothness assumptions,
then
the policy parameter~$\theta_t$ changes slowly enough to guarantee that
$G_t$ is a good approximation of the ideal $\nabla F_t(\theta)$ for OGD.
This, in turn, leads to the regret \eqref{eq:policy-regret} and local regret \eqref{eq:local-regret} guarantees of \GAPS~\citep{lin2024colt}.

These theoretical conditions naturally generalize a key property of linear dynamical systems under stabilizing policies.
However, they are conservative, and verification can be hard.
They are sufficient conditions for the local regret bound of \GAPS, but they are not known to be necessary.
Further, even when they are verified and the local regret bound \eqref{eq:local-regret} holds, it does not preclude the possibility of getting stuck in bad local minima.
We emphasize that \GAPS\ can be instantiated even when these assumptions cannot be proven.
It is important to validate its performance empirically in these cases.

\begin{algorithm}[t]
	\caption{Episodic, model-based (\DiffTune\ \citep{difftune})}
	\label{alg:pg-semimodel}
	\begin{algorithmic}
		\Require{parameter $\theta_1$, learning rate $\eta$, episode length $H$.}
		\For{each episode $k \in 1, \dots, \ceil{T/H}$}
			\State $y \gets \zero,\ \Gpg \gets \zero$.
			\For{$t \in (k-1)H, \dots, kH - 1$}
				\State Deploy $u_t = \pi_t(x_t, \theta_k)$ in environment.
				\State Observe $x_{t+1}$ and derivatives of $g_t$ and $f_t$.
				\State $\Gpg \gets \Gpg +
					(
						\partial f_t / \partial x_t +
						\partial f_t /\partial u_t \cdot \partial u_t / \partial x_t
					) y$
				\State $\hphantom{\Gpg \gets \Gpg} + \partial f_t / \partial u_t \cdot \partial u_t / \partial \theta_k$.
				\State Update $y$ as in \eqref{eq:mgaps-y} using parameter $\theta_k$.
			\EndFor
			\State $\theta_{k+1} \gets \theta_k - \eta \Gpg$.
		\EndFor
	\end{algorithmic}
\end{algorithm}

\subsection{Episodic methods}
\label{sec:baselines}

As discussed in the related work \S \ref{sec:related},
we compare \GAPS\ to episodic policy gradient methods but consider the episode length~$H$ a tunable hyperparameter in a non-episodic world.
For each episode $k \in 1, \dots, \ceil{T/H}$,
these methods deploy a constant parameter $\theta_k$ and approximate the gradient
\[
	\Gpg =
	\frac{\partial}{\partial \theta}
	\textstyle \sum_{t = (k - 1)H}^{kH - 1}
	f_t(x_t, u_t),
	\text{ assume }
	x_{(k-1)H} \text{ fixed}.
\]
We call this an approximation because it is computed 
as if the initial state $x_{(k-1)H}$ were fixed by nature,
whereas it actually depends on previously deployed parameters $\theta_1, \dots, \theta_{k-1}$.
We consider a
\emph{model-based} variant from the literature \citep{difftune} that maintains a sensitivity state recursively, similar to \GAPS,
and a \emph{model-free} variant that applies a state-of-the-art algorithm for derivative-free online optimization~\citep{zhang2024boosting}.

\subsubsection{Episodic, model-based (\DiffTune\ \citep{difftune})}
With known dynamics, we can compute $\Gpg$ analytically.
In contrast to model-based policy gradient methods that use a simulator in a sim-to-real paradigm, such as~\citep{mora2021pods},
we take the gradient about the real system's state trajectory and only use the model for derivatives.
We use the same sensitivity state~$y$ and recurrence \eqref{eq:mgaps-y} of \GAPS\ to compute the gradient efficiently,
yielding \Cref{alg:pg-semimodel}.
\citet{difftune} proposed this method and conducted hardware experiments similar to ours.
However, they treat episodes as part of ``nature'' and do not consider the effect of imposing episodes on a non-episodic problem.

{
\setlength{\textfloatsep}{0pt}
\begin{algorithm}[t]
	\caption{Episodic, model-free (\OnePoint\ \citep{zhang2024boosting})}
	\label{alg:pg-modelfree}
	\begin{algorithmic}
		\Require{parameter $\theta_1$, learning rate $\eta$, episode length $H$, perturbation radius $\epsilon$.}
		\State $J_0 \coloneqq 0$.
		\For{each episode $k \in 1, \dots, \ceil{T/H}$}
			\State $J_k \gets 0$.
			\State Sample perturbed ``query'' parameter:
			\State \quad \quad \quad
				$h_k \sim \cN(0, I),\ \theta'_k \gets \theta_{k-1} + \epsilon h_k$.
			\For{$t \in (k-1)H, \dots, kH - 1$}
				\State Deploy $u_t = \pi_t(x_t, \theta'_k)$ in environment.
				\State Observe $x_{t+1}$ and $c_t$.
				\State $J_k \gets J_k + c_t$.
			\EndFor
			\State $\theta_{k+1} \gets \theta_k - \frac{\eta}{\epsilon} (J_k - J_{k-1}) h_k$.
		\EndFor
	\end{algorithmic}
\end{algorithm}
}

\subsubsection{Episodic, model-free (\OnePoint\ \citep{zhang2024boosting})}
While \GAPS\ and \DiffTune\ use differentiable models,
it is also possible (and common in practice) to optimize policy parameters without knowing derivatives.
The most widely-used model-free policy gradient methods
use action-space randomness to estimate $\Gpg$ \citep{schulman-PPO}.
This is problematic for us because our quadrotor controller runs at \SI{500}{Hz}, far above the bandwidth of the physical system, so i.i.d. randomness at each step will be filtered away.
Instead,
we test a parameter-space policy gradient,
which was shown to compete well against action-space methods when the parameter $\theta$ is low-dimensional \citep{mania2018simple}.

We again apply ideas from online optimization for the time-varying setting.
In bandit online convex optimization with multiple queries allowed at each step, a two-point gradient method that queries at symmetrically perturbed parameters $\theta_k - h_k$, $\theta_k + h_k$, where $h_k$ is i.i.d. from a radially symmetric distribution, is essentially optimal 
\cite{agarwal2010optimal}.
However, when episodes are imposed on a single-trajectory setting,
it is not possible to evaluate the same episode cost twice.
Instead, we apply \emph{One-Point Residual Feedback}
(\OnePoint, \citep{zhang2024boosting}),
where past cost evaluations can reduce variance of single-point queries if costs change gradually.
In simulations \citep{zhang2024boosting,zhang2022onepoint},
\OnePoint\ improved substantially upon one-point methods without residual feedback.
We instantiate it for our setting in \Cref{alg:pg-modelfree}.

\section{Quadrotor Dynamics and Policy Class}
In this section we introduce our policy class and the quadrotor dynamics model used by \GAPS\ and \DiffTune.
Our policy follows a standard nonlinear geometric control structure 
with simplifications to speed up derivative computation.
We also propose reparameterizations of the state and policy to improve the optimization landscape.

\subsection{Dynamics and Representation}
Our dynamics model uses the Lie algebra $\so(3)$ of the special orthogonal group $\SO(3)$ to represent 3D rotation state.
Elements of $\so(3)$ are skew-symmetric matrices interpretable as 3D angular velocities.
We denote the natural isomorphism from $\R^3$ to $\so(3)$ by $\skewmat{\cdot}$:
if $x, y \in \R^3$, then $\skewmat{x} y = x \times y$, where $\times$ denotes the vector cross product.
The exponential map $\exp : \so(3) \mapsto \SO(3)$ gives the rotation reached after turning at the given angular velocity for one unit of time.
The logarithmic map $\log : \SO(n) \mapsto \so(n)$ is a continuous function satisfying $\exp(\log R) = R$, $\log(I_n) = \zero$.
\label{sec:representation}
The coordinates of $\so(3)$ provide a three-dimensional parameterization of $\SO(3)$,
making the quadrotor state space $\cX$ Euclidean instead of a manifold embedded in a higher-dimensional space.
This avoids the potential issue of the \GAPS\ state $y_t$ departing the tangent space of $\cX$ at $x_t$ due to compounded numerical errors.
The drawback is multiple-covering: the exponential map is many-to-one.
However, our experiments do not involve quadrotor flips and therefore will avoid this complication.\footnote{
    Representation of $\SO(3)$ by full matrices is the only way to avoid multiple covering, but applying \GAPS\ on the resulting manifold state space requires differential geometric considerations left for future work.
}

The quadrotor state is defined by
a position $p \in \R^3$ and velocity $v \in \R^3$ in the inertial frame,
a logarithmic coordinate
$\logR \in \so(3)$ such that $\exp(\logR)$ rotates from the body frame to the inertial frame,
and an angular velocity $\omega \in \so(3)$ in the body frame.
We consider an input space of $u = (\th, \tq)$,
where $\th \geq 0$ is a mass-normalized thrust
and $\tq = [\tq_r, \tq_p, \tq_y]^\top \in \R^3$ is the desired angular acceleration in the body frame.
We discretize the rotation dynamics with Lie group integration and the other states with forward Euler integration, yielding
\begin{gather}
\label{eq:dynamics-discrete}
	p_{t+1} = p_t + \dt v_t,
	\quad
	v_{t+1} = v_t + \dt (\th_t \exp(\logR_t) e_z - g ), \\
	\quad
	\logR_{t+1} = \log \left( \exp(\logR_t) \exp(\dt \omega_t) \right),
	\quad
	\omega_{t+1} = \omega_t + \dt \tq, \notag
\end{gather}
where $\dt > 0$ is the discretization time interval and $g$ is the gravitational constant.
Our model is equivalent to having unit mass and identity moment of inertia.
For simplicity, we assume that a lower-level thrust allocator can realize $u$ based on known mass, inertia, motor, and propeller characteristics.

\emph{Target trajectory and error state.}
Let
$p^d : \R_{\geq 0} \mapsto \R^3$
denote a thrice-differentiable target position trajectory, with the corresponding target velocity and acceleration denoted by 
$v^d(t) = \dot p^d(t)$
and
$a^d(t) = \ddot p^d(t)$.
To express controllers with integral action, we add the virtual state $\ierr \in \R^3$
with dynamics
\[
    \ierr_0 = \zero,\quad \ierr_{t+1} = \ierr_t + \dt (p_t - p^d_t)\ \forall t \geq 0,
\]
and define the full state $x_t = (\ierr_t, p_t, v_t, r_t, \omega_t)$.
By the quadrotor's differential flatness \citep{mellinger-minsnap},
one can compute a desired body-frame angular velocity~$\omega^d_t$ from higher-order derivatives of $p^d$.

\subsection{Policy class}
Our policy class follows a widely-used structure for quadrotor trajectory-tracking controllers \cite{lee2010geometric,mellinger-minsnap}.
An outer loop of position control calculates a desired mass-normalized thrust vector~$\tdes \in \R^3$,
and an inner loop of geometric attitude control orients the quadrotor's body towards $\tdes$.
The outer-loop law is
\begin{equation}
\label{eq:ctrl-position}
    \tdes = 
    - K_i \ierr
    - K_p (p - p^d)
    - K_v (v - v^d)
    + a^d
    + g e_z,
\end{equation}
where $K_i, K_p, K_v$ are positive-definite diagonal gain matrices.
The scalar thrust command $\th$ is computed by projecting $\tdes$ onto the body thrust axis: $\th = \tdes^\top \exp(\logR)e_z$.
The inner loop begins by constructing a desired attitude $r_d \in \so(3)$
as the shortest rotation that takes $e_z$ to $\tdes$, given by
\begin{equation*}
\label{eq:ctrl-attitude-Rdes}
    r_d \coloneqq
    \cos^{-1}\left(\nofrac{e_z \cdot \tdes}{\norm{\tdes}}\right)
    \skewmat{ \nofrac{e_z \times \tdes}{\norm{e_z \times \tdes}} },
\end{equation*}
or $r_d = \zero$ when $e_z \times \tdes = \zero$,
and $\norm{z} > 0$ by assumption that the acceleration and error terms in \eqref{eq:ctrl-position} are sufficiently small.
Unlike many quadrotor controllers, ours has no concept of desired heading (yaw).
The yaw portion of $r_d$ will be near zero when $\tdes$ is near $e_z$.
This simplifies the derivatives needed by \GAPS\ considerably and is adequate for our experiments, but a desired yaw could be added.
The desired angular acceleration $\tau$ is determined by
\begin{align*}
\label{eq:ctrl-attitude-gain}
    \tq' &\coloneqq
        - K_r \log(\exp(\logR) \exp(-\logR_d))
        - K_\omega (\omega - \omega^d), \\
    \tq &= \softclamp(\tq', [B_{xy}, B_{xy}, B_z]^\top),
\end{align*}
where $K_r, K_\omega$ are positive-definite diagonal gain matrices,
$\softclamp(x, y) = y \tanh(x/y)$ is applied elementwise,
and $B_{xy}, B_z > 0$ are user-defined upper bounds.
The $\softclamp$ is critical for high-gain attitude control using this cascaded structure.
Without it, large attitude errors may lead to commands that cannot be realized without setting some of the motor thrusts to zero, causing instability and altitude loss.

\figbadinitfigeight

\subsection{Contractiveness}
A geometric controller similar to ours 
was shown to be exponentially stabilizing when the attitude error
is less than 90 degrees \citep{lee2010geometric}.
In Euclidean state spaces, exponential stability for tracking controllers is generally sufficient for contractiveness in the sense required by \GAPS.
However, the analysis of \cite{lee2010geometric} considers exponential stability directly on the $\SO(3)$ manifold, whereas we use the logarithmic parameterization of rotations to avoid a manifold state space.
Deriving contractiveness of $\logR$ from exponential stability of~$\exp(\logR)$ requires differential geometry arguments beyond the scope of this work.
Also, the analysis of \cite{lee2010geometric} does not account for actuation limits, which we found to be critical and addressed by the $\softclamp$ in our design.
Still, we conjecture that with suitable assumptions on the initial error and disturbances, our controller and state parameterization is contractive.

\subsection{Logarithmic Policy Reparameterization}
\label{sec:log-coords}
To improve the optimization landscape conditioning, 
we use a logarithmic reparameterization of the feedback gains.
In our policy class, $K_{i,p,v,r,\omega}$ are nonnegative diagonal matrices
whose entries can range over several orders of magnitude.
In the expert-tuned baseline $\thetamanual$ discussed in \S \ref{sec:crazyflie}, 
the largest gain is $\approx 1000$, while the smallest is $\approx 1$.
Further, because the elements of $\theta$ have highly distinct physical meanings, the sensitivity of the surrogate cost $F_t$ is nonuniform:
from empirical observations, we have that $\abs{\nabla F_t}$ with respect to each entry $k$ is roughly proportional to $1/\abs{k}$.
To handle this simply, we let $\theta$ determine the \emph{logarithms} of the gains.
Due to the geometric symmetries of the quadrotor,
we use equal gains for the two horizontal axes, e.g. $K_r = \diag(\krxy, \krxy, \krz)$.
This yields the ``raw'' parameter
\[
    \vartheta = (\kixy, \kiz, \kpxy, \kpz, \kvxy, \kvz, \krxy, \krz, \kwxy, \kwz) \in \R^{10}_{> 0}
\]
and the reparameterization
$\theta = \log(\vartheta)$,
which we optimize with \GAPS, \DiffTune, and \OnePoint.
We fix the cost function for all experiments and all timesteps $t$ as
\begin{align*}
    f_t(x, u) = {} & \dt \Big(
    \norm{p - p^d}_2^2
    + 10^{-4} \norm{v - v^d}_2^2 \\
    &+ 10^{-3} \norm{\omega - \omega^d}_2^2
    + 10^{-7} \norm{\tq}_2^2
    + 10^{-8} \th^2 \Big).
\end{align*}
This mainly penalizes position tracking error, with other terms acting to regularize against high control effort.
Regularization constants were chosen empirically to be as small as possible while preventing any tendency towards attitude oscillations.

\section{Quadrotor Experiments}
\label{sec:crazyflie}

Our quadrotor is a Bitcraze Crazyflie 2.0 with the manufacturer's thrust upgrade bundle and a larger battery.
For the analytic derivatives needed by \GAPS\ and \DiffTune\ (\ref{eq:mgaps-y}-\ref{eq:mgaps-grad}),
we use the SymForce package~\citep{martiros22symforce},
which includes symbolic differentiation of Lie algebraic operations
and generates C++ code suitable for embedded systems.
All algorithms run at \SI{500}{Hz} on the Crazyflie 2.0's \SI{168}{MHz} onboard microcontroller with \SI{192}{kB} memory.
The controlling PC runs the Crazyswarm package~\citep{preiss2017icra}.
\textbf{As a baseline for comparison, we have access to expert controller parameters $\thetamanual$ manually tuned over several days of effort~\citep{preiss2017icra}.}

We start with a trajectory tracking experiment where the disturbances are small (Subsections A-B),
and test how quickly each policy optimization algorithm approaches the near-optimal behavior of $\thetamanual$ when initialized with a suboptimal parameter.
Then, we experiment with larger disturbances from time-varying wind or a heavy payload (Subsection C), which $\thetamanual$ is not tuned to handle.
If \GAPS\ outperforms $\thetamanual$ in such settings, it suggests that no fixed parameter is near-optimal in all circumstances, motivating online policy optimization.
When applicable, all figures use shorthand names for optimizers and fixed parameter values given in \Cref{tab:optimizers}.

\begin{table}[htpb]
    \centering
    \caption{
        Shorthand names for optimizers and parameter values
        used in Figures \ref{fig:bad-init-fig8}-\ref{fig:fan-error}.
    }
    \label{tab:optimizers}
    \begin{tabular}{ll}
        \toprule
        \textbf{Name} & \textbf{Description} \\
        \midrule
        \texttt{Expert} & $\theta_t = \thetamanual \ \forall t$.
            \hspace{5.6mm} $\thetamanual$ = Manually tuned, near-optimal. \\
        \texttt{Detune} & $\theta_t = \thetamanual - \log 2 \ \forall t$.
            \; Stabilizing but far from optimal. \\
        \GAPS & \Cref{alg:mgaps}. \\
        \DiffTune$^\star$ & \Cref{alg:pg-semimodel} with hindsight best episode length. \\
        \DiffTune & \Cref{alg:pg-semimodel} with hindsight worst episode length. \\
        \OnePoint & \Cref{alg:pg-modelfree} using \DiffTune$^\star$'s episode length. \\
        \bottomrule
    \end{tabular}
\end{table}

\subsection{Detuned initialization}
\label{sec:bad-init}

We simulate the process of controller tuning with a new robot or policy class
by initializing each algorithm with the ``detuned'' parameter ${\theta_0 = \thetamanual - \log 2}$,
which corresponds with halving the feedback gains due to our logarithmic policy reparameterization.
This gives a policy that stabilizes near hover, but is less suited for aggressive trajectories.
We deploy \GAPS, \DiffTune, and \OnePoint\
in an aggressive figure-8 trajectory moving in a diagonal plane, such that high thrust and roll/pitch torques are needed.

The trajectories for the expert $\thetamanual$, detuned $\theta_0$, and each of the optimization algorithms are shown \Cref{fig:bad-init-fig8}.
To illustrate the impact of episode length, \DiffTune\ is shown with two different episode lengths, with \emph{\DiffTune$^\star$} indicating the best-performing length in hindsight, and \emph{\DiffTune} the worst (details in \S \ref{sec:episode-length}).
Color corresponds to time.
The expert and detuned trajectories are nearly unchanging, as expected.
In contrast, all algorithms significantly alter the trajectory.
\GAPS\ is initially near \emph{detune}, but after 2-3 laps (8-12 sec) is near \emph{expert}.
\DiffTune\ with optimal episode length does not improve on the first lap but jumps near \emph{expert} after its first update,
while the improvement is much slower with a suboptimal episode length.
\OnePoint\ improves some, but is much slower than the model-based methods.

\figbadinitcost

The improvement is shown quantitatively in \Cref{fig:bad-init-cost}.
We are interested in the policy regret \eqref{eq:policy-regret},
but since the true clairvoyant optimal $\theta$ is unknown, we use the manually-tuned $\thetamanual$ as a surrogate,
and plot the cumulative cost difference from $\thetamanual$ for each method,
henceforth called \emph{quasi-regret}.
This scenario is disturbance-free, so $\thetamanual$ is near-optimal.
We see \GAPS\ quickly reduces cost and converges to constant quasi-regret against $\thetamanual$.
\DiffTune\ with optimal episode length converges to a larger constant quasi-regret.
However, with a bad episode length its quasi-regret is significantly degraded and still growing as the experiment ends.
The model-free \OnePoint\ has substantially higher quasi-regret.

\figbadinitparams

The evolution of parameters for each algorithm are shown in \Cref{fig:bad-init-params}.
The model-free \OnePoint\ produces noisy parameter changes due to its randomized queries and high-variance gradient estimate.
The model-based episodic gradient of \DiffTune\ behaves more similarly to \GAPS, but illustrates the additional drawback of large steps in the policy parameters.
The large steps caused jitter in the quadrotor's attitude, which was visible in person but is not clear in the position trajectory or quasi-regret plots.

\subsection{\DiffTune\ sensitivity to episode length}

\label{sec:episode-length}
In \Cref{fig:episodic},
we expand the comparison of \GAPS\ and \DiffTune.
Total costs
in the detuned initialization scenario
are shown for \GAPS\ and
\DiffTune\ with various episode lengths.
The figure-8 trajectory lasts \SI{4}{sec} and is symmetric.
Therefore, with our \SI{500}{Hz} controller, episodes of $1000$ and $2000$ are especially well-aligned,
as the data confirms.
In real-world applications that are not periodic, the optimal episode length may not be so obvious, and might also change over time.
Episode length becomes another hyperparameter to empirically tune.
In contrast, learning rate is the only hyperparameter of \GAPS.

\figepisodic

\subsection{Large-Disturbance Experiments}

\figfancost

\subsubsection{Time-varying wind}
\label{sec:fan}
\figfanparams
To study short-term adaptivity,
we create a periodic wind disturbance with three household box fans side-by-side.
To simulate the powerful wind disturbances of outdoor flight, we attach a cardboard panel to the quadrotor that presents a large ``sail'' to the wind, magnifying its effect.
We fly a linear back-and-forth pattern in front of the fan array, with each ``lap'' taking \SI{4}{sec}.
The fan power is toggled every 3 laps (\SI{12}{sec}).
Tracking error is shown in \Cref{fig:fan-error}.
We see that \GAPS\ substantially improves tracking accuracy in both fan-on (shaded) and fan-off phases.
The evolution of each parameter is plotted in \Cref{fig:fan-params}.
Several parameters show significant transient responses when the phase switches.
In particular, $\kpxy$ approaches oscillation.
This confirms that \GAPS\ adapts quickly to scenario changes on the scale of ten seconds.
The dynamics model used by \GAPS\ is unchanged from nominal,
demonstrating robustness against model error.

\figweightcost

\subsubsection{Heavy payload}
\label{sec:weight}
To further study robustness against modeling error, we attach a heavy weight to the quadrotor while leaving the dynamics model used by \GAPS\ unchanged from the nominal mass.
Our quadrotor's mass is \SI{39}{g}.
We attach a \SI{23}{g} steel weight near its center, increasing mass by \SI{60}{\%} and acting as a large, near-constant disturbance force.
This is near the quadrotor's thrust limit, so we fly in a horizontally-oriented circle that does not require excess thrust.
The parameter $\thetamanual$ was not tuned to handle such large disturbances.
In \Cref{fig:weight},
we observe that \GAPS\ is able to substantially reduce the tracking error compared to $\thetamanual$.

\figcompareparams

\subsubsection{Parameter comparison between disturbance types}
In \Cref{fig:compare-params}, we compare the parameter evolution for the time-varying-wind and heavy-payload  scenarios.
We observe large differences.
In particular, $\kpz$ has the largest change in the heavy-payload scenario, but changes minimally in the wind scenario.
The weight acts in the $z$-axis while the fan acts in the \mbox{$y$-axis}.
This confirms that \GAPS\ is adapting to the specific disturbances experienced in each scenario, rather than (or in addition to) the intrinsic properties of the quadrotor.

\section{Ackermann-Steered Car Experiment}
\label{sec:traxxas}

To evaluate the versatility of \GAPS\ with respect to robot type,
we deploy it on an Ackermann-steered 1:6-scale remote-control car (Traxxas X-Maxx)
with onboard visual-inertial localization,
driving in a circle on grass and concrete.
Details of the state representation, dynamics model, policy class, and cost function are given in \Cref{appx:traxxas}.
We use \GAPS\ to tune the controller after finding a suboptimal stabilizing controller with limited manual effort.
\GAPS\ optimizes squared tracking error with small angular velocity error and steering angle regularization terms.
Results are shown in \Cref{fig:traxxas}.
\GAPS\ reduces the tracking error by a factor of around $5$ within \SI{10}{sec} of being enabled -- less than one full circle period.
This shows that 1) the positive results of \GAPS\ are not restricted to quadrotors, and 2) \GAPS\ can be useful for automatic tuning on new hardware or a new controller design, even if online adaptivity is not the main goal.

\figtraxxas

\section{Conclusion}
We experimentally evaluated algorithms that apply the principle of online gradient descent to the online policy optimization problem for nonlinear robot controllers.
For tuning a geometric quadrotor controller for aggressive flight from a suboptimal initialization,
the non-episodic model-based \GAPS\ performed best, finding a near-optimal policy in about \SI{15}{sec}.
The model-based episodic \DiffTune\ performed nearly as well when its episode length was ideal, but degraded with other episode lengths.
The model-free episodic \OnePoint\ further lagged the model-based methods.
We also validated \GAPS\ in a similar bad-initialization experiment on a small off-road car, showing versatility with respect to robot platforms.

To further evaluate \GAPS, we compared it to the expert-tuned parameter $\thetamanual$ under disturbances from a heavy weight and periodic wind.
In both scenarios, \GAPS\ improved substantially from $\thetamanual$ while finding very different optimized parameters.
This shows that \GAPS\ adapts to the specific problem instance, and can improve performance even when the dynamics model is not perfect.
More broadly, our results show the benefit of online adaptivity in general:
by allowing the core controller parameters to change moment-to-moment in response to current real-world conditions,
we achieve strong performance in a wide range of scenarios.

\section{Limitations}
The requirement of a differentiable dynamics may limit the applicability of \GAPS\ and \DiffTune\ to systems with discontinuities, such as walking robots.
Unstable behavior near discontinuities can degrade gradient-based optimization, even when the discontinuities form a set of measure zero \cite{suh2022differentiable}.
Further testing and development is needed on such systems.

The contractiveness required for the local regret guarantee of \GAPS\ can be difficult to verify,
and does not rule out the possibility of bad local minima.
Similar to other applications of gradient-based nonconvex optimization,
one must do empirical validation (like this work) before deploying \GAPS\ in complex real-world robotic systems.

\section*{Acknowledgments}
The authors thank E.-S. Lupu for stimulating discussion and assistance with the car experiment.

\bibliographystyle{plainnat}
\bibliography{references}

\begin{thebibliography}{42}
\providecommand{\natexlab}[1]{#1}
\providecommand{\url}[1]{\texttt{#1}}
\expandafter\ifx\csname urlstyle\endcsname\relax
  \providecommand{\doi}[1]{doi: #1}\else
  \providecommand{\doi}{doi: \begingroup \urlstyle{rm}\Url}\fi

\bibitem[Agarwal et~al.(2010)Agarwal, Dekel, and Xiao]{agarwal2010optimal}
Alekh Agarwal, Ofer Dekel, and Lin Xiao.
\newblock
  \href{http://colt2010.haifa.il.ibm.com/papers/COLT2010proceedings.pdf\#page=36}{Optimal
  Algorithms for Online Convex Optimization with Multi-Point Bandit Feedback}.
\newblock In \emph{\COLT}, 2010.

\bibitem[Agarwal et~al.(2019)Agarwal, Bullins, Hazan, Kakade, and
  Singh]{agarwal2019online}
Naman Agarwal, Brian Bullins, Elad Hazan, Sham~M. Kakade, and Karan Singh.
\newblock \href{http://proceedings.mlr.press/v97/agarwal19c.html}{Online
  Control with Adversarial Disturbances}.
\newblock In \emph{\ICML}, 2019.

\bibitem[Amos et~al.(2018)Amos, Rodriguez, Sacks, Boots, and
  Kolter]{amos2018DiffMPC}
Brandon Amos, Ivan Dario~Jimenez Rodriguez, Jacob Sacks, Byron Boots, and
  J.~Zico Kolter.
\newblock
  \href{https://proceedings.neurips.cc/paper/2018/hash/ba6d843eb4251a4526ce65d1807a9309-Abstract.html}{Differentiable
  {MPC} for End-to-end Planning and Control}.
\newblock In \emph{\NeurIPS}, 2018.

\bibitem[Amos et~al.(2021)Amos, Stanton, Yarats, and
  Wilson]{amos2021modelbased}
Brandon Amos, Samuel Stanton, Denis Yarats, and Andrew~Gordon Wilson.
\newblock On the model-based stochastic value gradient for continuous
  reinforcement learning.
\newblock In \emph{{\LFOURDC}}, 2021.

\bibitem[Baxter and Bartlett(2001)]{baxter2001infinite}
Jonathan Baxter and Peter~L Bartlett.
\newblock Infinite-horizon policy-gradient estimation.
\newblock \emph{Journal of Artificial Intelligence Research}, 15:\penalty0
  319--350, 2001.

\bibitem[Bristow et~al.(2006)Bristow, Tharayil, and Alleyne]{bristow2006ILC}
Douglas~A Bristow, Marina Tharayil, and Andrew~G Alleyne.
\newblock A survey of iterative learning control.
\newblock \emph{IEEE Control Systems Magazine}, 26\penalty0 (3):\penalty0
  96--114, 2006.

\bibitem[Cheng et~al.(2024)Cheng, Kim, Song, Yang, Jin, Wang, and
  Hovakimyan]{difftune}
Sheng Cheng, Minkyung Kim, Lin Song, Chengyu Yang, Yiquan Jin, Shenlong Wang,
  and Naira Hovakimyan.
\newblock Difftune: Autotuning through autodifferentiation.
\newblock \emph{\TRO}, 40:\penalty0 4085--4101, 2024.

\bibitem[Coros et~al.(2021)Coros, Macklin, Thomaszewski, and
  Th{\"u}rey]{coros2021differentiable}
Stelian Coros, Miles Macklin, Bernhard Thomaszewski, and Nils Th{\"u}rey.
\newblock Differentiable simulation.
\newblock In \emph{SIGGRAPH Asia 2021 Courses}, pages 1--142. 2021.

\bibitem[de~Avila Belbute-Peres et~al.(2018)de~Avila Belbute-Peres, Smith,
  Allen, Tenenbaum, and Kolter]{deavila2018diffphysics}
Filipe de~Avila Belbute-Peres, Kevin Smith, Kelsey Allen, Josh Tenenbaum, and
  J.~Zico Kolter.
\newblock
  \href{https://proceedings.neurips.cc/paper_files/paper/2018/file/842424a1d0595b76ec4fa03c46e8d755-Paper.pdf}{End-to-End
  Differentiable Physics for Learning and Control}.
\newblock In \emph{\NeurIPS}, 2018.

\bibitem[Gradu et~al.(2023)Gradu, Hazan, and Minasyan]{gradu2023adaptive}
Paula Gradu, Elad Hazan, and Edgar Minasyan.
\newblock \href{https://proceedings.mlr.press/v211/gradu23a.html}{Adaptive
  Regret for Control of Time-Varying Dynamics}.
\newblock In \emph{\LFOURDC}, 2023.

\bibitem[Haarnoja et~al.(2018)Haarnoja, Zhou, Abbeel, and Levine]{haarnoja-sac}
Tuomas Haarnoja, Aurick Zhou, Pieter Abbeel, and Sergey Levine.
\newblock Soft actor-critic: Off-policy maximum entropy deep reinforcement
  learning with a stochastic actor.
\newblock In \emph{\ICML}, 2018.

\bibitem[Hazan(2022)]{hazan2022oco}
Elad Hazan.
\newblock \emph{Introduction to Online Convex Optimization}.
\newblock MIT Press, 2022.

\bibitem[Hazan and Singh(2022)]{hazan2022nonstochasticbook}
Elad Hazan and Karan Singh.
\newblock \href{https://doi.org/10.48550/arXiv.2211.09619}{Introduction to
  Online Nonstochastic Control}.
\newblock \emph{CoRR}, abs/2211.09619, 2022.

\bibitem[Hazan et~al.(2020)Hazan, Kakade, and Singh]{hazan2020nonstochastic}
Elad Hazan, Sham Kakade, and Karan Singh.
\newblock \href{https://proceedings.mlr.press/v117/hazan20a.html}{The
  Nonstochastic Control Problem}.
\newblock In \emph{\ALT}, 2020.

\bibitem[Ioannou and Sun(1995)]{ioannou1995robust}
Petros~A. Ioannou and Jing Sun.
\newblock \emph{Robust Adaptive Control}.
\newblock Prentice-Hall, 1995.

\bibitem[Kalashnikov et~al.(2018)Kalashnikov, Irpan, Pastor, Ibarz, Herzog,
  Jang, Quillen, Holly, Kalakrishnan, Vanhoucke, et~al.]{kalashnikov2018qtopt}
Dmitry Kalashnikov, Alex Irpan, Peter Pastor, Julian Ibarz, Alexander Herzog,
  Eric Jang, Deirdre Quillen, Ethan Holly, Mrinal Kalakrishnan, Vincent
  Vanhoucke, et~al.
\newblock Scalable deep reinforcement learning for vision-based robotic
  manipulation.
\newblock In \emph{{\CORL}}, 2018.

\bibitem[Krstic et~al.(1995)Krstic, Kokotovic, and
  Kanellakopoulos]{krstic1995nonlinear}
Miroslav Krstic, Petar~V. Kokotovic, and Ioannis Kanellakopoulos.
\newblock \emph{Nonlinear and Adaptive Control Design}.
\newblock John Wiley \& Sons, 1995.

\bibitem[Lee et~al.(2010)Lee, Leok, and McClamroch]{lee2010geometric}
Taeyoung Lee, Melvin Leok, and N.~Harris McClamroch.
\newblock \href{https://doi.org/10.1109/CDC.2010.5717652}{Geometric tracking
  control of a quadrotor {UAV} on {SE(3)}}.
\newblock In \emph{\CDC}, 2010.

\bibitem[Lin et~al.(2023)Lin, Preiss, Anand, Li, Yue, and Wierman]{lin2023gaps}
Yiheng Lin, James~A. Preiss, Emile Anand, Yingying Li, Yisong Yue, and Adam
  Wierman.
\newblock Online adaptive policy selection in time-varying systems: No-regret
  via contractive perturbations.
\newblock In \emph{\NeurIPS}, 2023.

\bibitem[Lin et~al.(2024)Lin, Preiss, Xie, Anand, Chung, Yue, and
  Wierman]{lin2024colt}
Yiheng Lin, James~A. Preiss, Fengze Xie, Emile Anand, Soon-Jo Chung, Yisong
  Yue, and Adam Wierman.
\newblock Online policy optimization in unknown nonlinear systems.
\newblock In \emph{{\COLT}}, 2024.

\bibitem[Lopez and Slotine(2021)]{lopez2021universal}
Brett~T Lopez and Jean-Jacques~E Slotine.
\newblock Universal adaptive control of nonlinear systems.
\newblock \emph{IEEE Control Systems Letters}, 6:\penalty0 1826--1830, 2021.

\bibitem[Lupu et~al.(2024)Lupu, Xie, Preiss, Alindogan, Anderson, and
  Chung]{magicvfm}
Elena~Sorina Lupu, Fengze Xie, James~A. Preiss, Jedidiah Alindogan, Matthew
  Anderson, and Soon-Jo Chung.
\newblock {MAGIC-VFM}: Meta-learning adaptation for ground interaction control
  with visual foundation models.
\newblock \emph{\TRO}, pages 1--20, 2024.

\bibitem[Mania et~al.(2018)Mania, Guy, and Recht]{mania2018simple}
Horia Mania, Aurelia Guy, and Benjamin Recht.
\newblock Simple random search of static linear policies is competitive for
  reinforcement learning.
\newblock In \emph{{\NeurIPS}}, 2018.

\bibitem[Martiros et~al.(2022)Martiros, Miller, Bucki, Solliday, Kennedy, Zhu,
  Dang, Pattison, Zheng, Tomic, Henry, Cross, VanderMey, Sun, Wang, and
  Holtz]{martiros22symforce}
Hayk Martiros, Aaron Miller, Nathan Bucki, Bradley Solliday, Ryan Kennedy, Jack
  Zhu, Tung Dang, Dominic Pattison, Harrison Zheng, Teo Tomic, Peter Henry,
  Gareth Cross, Josiah VanderMey, Alvin Sun, Samuel Wang, and Kristen Holtz.
\newblock {SymForce: Symbolic Computation and Code Generation for Robotics}.
\newblock In \emph{{\RSS}}, 2022.

\bibitem[Mellinger and Kumar(2011)]{mellinger-minsnap}
Daniel Mellinger and Vijay Kumar.
\newblock Minimum snap trajectory generation and control for quadrotors.
\newblock In \emph{{\ICRA}}, 2011.

\bibitem[Mo and Farid(2019)]{mo2019nonlinear}
Hongwei Mo and Ghulam Farid.
\newblock Nonlinear and adaptive intelligent control techniques for quadrotor
  {UAV}--a survey.
\newblock \emph{Asian Journal of Control}, 21\penalty0 (2):\penalty0 989--1008,
  2019.

\bibitem[Mora et~al.(2021)Mora, Peychev, Ha, Vechev, and Coros]{mora2021pods}
Miguel Angel~Zamora Mora, Momchil Peychev, Sehoon Ha, Martin Vechev, and
  Stelian Coros.
\newblock Pods: Policy optimization via differentiable simulation.
\newblock In \emph{{\ICML}}, 2021.

\bibitem[O’Connell et~al.(2022)O’Connell, Shi, Shi, Azizzadenesheli,
  Anandkumar, Yue, and Chung]{neuralfly}
Michael O’Connell, Guanya Shi, Xichen Shi, Kamyar Azizzadenesheli, Anima
  Anandkumar, Yisong Yue, and Soon-Jo Chung.
\newblock Neural-fly enables rapid learning for agile flight in strong winds.
\newblock \emph{Science Robotics}, 7\penalty0 (66), 2022.

\bibitem[Paternain et~al.(2022)Paternain, Bazerque, and
  Ribeiro]{paternain2022continuing}
Santiago Paternain, Juan~Andr{\'{e}}s Bazerque, and Alejandro Ribeiro.
\newblock Policy gradient for continuing tasks in discounted {Markov} decision
  processes.
\newblock \emph{{\TAC}}, 67\penalty0 (9):\penalty0 4467--4482, 2022.

\bibitem[Preiss et~al.(2017)Preiss, H{\"{o}}nig, Sukhatme, and
  Ayanian]{preiss2017icra}
James~A. Preiss, Wolfgang H{\"{o}}nig, Gaurav~S. Sukhatme, and Nora Ayanian.
\newblock Crazyswarm: {A} large nano-quadcopter swarm.
\newblock In \emph{\ICRA}, 2017.

\bibitem[Schenck and Fox(2018)]{schenck2018differentiable}
Connor Schenck and Dieter Fox.
\newblock \href{https://proceedings.mlr.press/v87/schenck18a.html}{SPNets:
  Differentiable Fluid Dynamics for Deep Neural Networks}.
\newblock In \emph{\CORL}, 2018.

\bibitem[Schulman et~al.(2017)Schulman, Wolski, Dhariwal, Radford, and
  Klimov]{schulman-PPO}
John Schulman, Filip Wolski, Prafulla Dhariwal, Alec Radford, and Oleg Klimov.
\newblock Proximal policy optimization algorithms.
\newblock \emph{CoRR}, abs/1707.06347, 2017.

\bibitem[Sehnke et~al.(2010)Sehnke, Osendorfer, R{\"u}ckstie{\ss}, Graves,
  Peters, and Schmidhuber]{sehnke2010parameter}
Frank Sehnke, Christian Osendorfer, Thomas R{\"u}ckstie{\ss}, Alex Graves, Jan
  Peters, and J{\"u}rgen Schmidhuber.
\newblock Parameter-exploring policy gradients.
\newblock \emph{Neural Networks}, 23\penalty0 (4):\penalty0 551--559, 2010.

\bibitem[Silver et~al.(2014)Silver, Lever, Heess, Degris, Wierstra, and
  Riedmiller]{silver2014deterministic}
David Silver, Guy Lever, Nicolas Heess, Thomas Degris, Daan Wierstra, and
  Martin Riedmiller.
\newblock Deterministic policy gradient algorithms.
\newblock In \emph{{\ICML}}, 2014.

\bibitem[Simchowitz et~al.(2020)Simchowitz, Singh, and
  Hazan]{simchowitz2020improper}
Max Simchowitz, Karan Singh, and Elad Hazan.
\newblock \href{https://proceedings.mlr.press/v125/simchowitz20a.html}{Improper
  Learning for Non-Stochastic Control}.
\newblock In \emph{\COLT}, 2020.

\bibitem[Slotine and Li(1991)]{slotine1991applied}
J.J.E. Slotine and W.~Li.
\newblock \emph{Applied Nonlinear Control}.
\newblock Prentice Hall, 1991.

\bibitem[Suh et~al.(2022)Suh, Simchowitz, Zhang, and
  Tedrake]{suh2022differentiable}
Hyung~Ju Suh, Max Simchowitz, Kaiqing Zhang, and Russ Tedrake.
\newblock Do differentiable simulators give better policy gradients?
\newblock In \emph{{\ICML}}, 2022.

\bibitem[Werbos(1991)]{werbos1991menu}
Paul~J. Werbos.
\newblock {A Menu of Designs for Reinforcement Learning Over Time}.
\newblock In \emph{{Neural Networks for Control}}. MIT Press, 1991.

\bibitem[Williams(1992)]{williams-REINFORCE}
Ronald~J. Williams.
\newblock Simple statistical gradient-following algorithms for connectionist
  reinforcement learning.
\newblock \emph{Machine Learning}, 8:\penalty0 229--256, 1992.

\bibitem[Xue et~al.(2023)Xue, Liao, Gan, Park, Xie, Liu, and
  Cao]{xue2023JAXFEM}
Tianju Xue, Shuheng Liao, Zhengtao Gan, Chanwook Park, Xiaoyu Xie, Wing~Kam
  Liu, and Jian Cao.
\newblock \href{https://doi.org/10.1016/j.cpc.2023.108802}{{JAX-FEM:} {A}
  differentiable {GPU}-accelerated {3D} finite element solver for automatic
  inverse design and mechanistic data science}.
\newblock \emph{Comput. Phys. Commun.}, 291:\penalty0 108802, 2023.

\bibitem[Zhang et~al.(2022)Zhang, Zhou, Ji, and Zavlanos]{zhang2022onepoint}
Yan Zhang, Yi~Zhou, Kaiyi Ji, and Michael~M. Zavlanos.
\newblock \href{https://dblp.org/rec/journals/automatica/ZhangZJZ22.bib}{A new
  one-point residual-feedback oracle for black-box learning and control}.
\newblock \emph{Automatica}, 136:\penalty0 110006, 2022.

\bibitem[Zhang et~al.(2024)Zhang, Zhou, Ji, Shen, and
  Zavlanos]{zhang2024boosting}
Yan Zhang, Yi~Zhou, Kaiyi Ji, Yi~Shen, and Michael~M. Zavlanos.
\newblock Boosting one-point derivative-free online optimization via residual
  feedback.
\newblock \emph{{\TAC}}, pages 1--8, 2024.

\end{thebibliography}

\newpage
\appendix
\subsection{Details of Ackermann-Steered Car Experiment}
\label{appx:traxxas}

In this appendix we give details of the 1:6-scale car experiment from \Cref{sec:traxxas}.

\subsubsection{Dynamics and Representation}
We consider a planar dynamics model.
The system state comprises the car's
position ${p \in \mathbb{R}^2}$,
body-frame velocity $v \in \mathbb{R}^2$,
heading angle $r \in \so(2)$,
angular velocity $\omega \in \so(2)$,
and steering angle $\psi \in \R$.
(Note that we overload notations for analogous states with the quadrotor.)
The velocity components $v = (v^x, v^y)$ correspond to the forward and left directions respectively.
The heading~$r$ rotates from the body frame to the inertial frame and is parameterized by the Lie algebra $\so(2)$.
The control input $u = (\th, \tq) \in \R^2$
contains the throttle and steering inputs respectively.
We denote the lateral dynamics velocity states by $s = (v^y, \omega)$.
The dynamics take the control-affine form
\begin{align*}
    p_{t+1} &= p_t + \delta \exp(r_t)v_t, \\
    r_{t+1} &= \log(\exp(r_t)\exp(\delta\omega_t)), \\
    \quad s_{t + 1} &= s_t + \delta A(v^x_t)s_t + \delta B \psi_t, \\
    v^x_{t+1} &= v^x_t + \delta C_\th (\th_t - v^x_t), \\
    \psi_{t+1} &= \psi_t + \delta C_\tq (\tq_t - \psi_t),
\end{align*}
where $\delta > 0$ is the discretization interval,
$0 < C_\th, C_\tq \ll 1/\delta$ are time delay constants,
and the function $A : \R \mapsto \R^{2 \times 2}$ and matrix $B \in \R^{2 \times 2}$ are given by~\cite{magicvfm}.
A smooth desired trajectory $p^d : \R_{\geq 0} \mapsto \R^2$ is specified,
and we compute the corresponding desired heading and body-frame linear velocity
\[
	r^d = \operatorname{atan2}(\dot p^d_y, \dot p^d_x),
	\quad
	v^d = \exp(-r^d) \dot p^d,
\]
and the desired angular velocity
\[
	\omega^d =
	\dot r^d =
	\recip{\norm{\dot p^d}_2^2} \left(
		\dot p^d_x \ddot p^d_y
		- \dot p^d_y \ddot p^d_x
	\right),
\]
where we have used the chain rule and assumed that $\dot p^d \neq \zero$ throughout the trajectory.

\subsubsection{Policy class}
The policy class is:
\begin{align*}
    \tq_t =&
        -K_1  \left[ \exp(-r^d_t) (p_t - p^d_t) \right]_{y}  \\
        & -K_2 (v^y_t + r^e_t v^x_t)
        -K_3 r^e_t
        -K_4 (\omega_t - \omega^d_t), \\[2mm]
    \th_t =&
        -K_p \left[ \exp(-r_t) (p_t - p^d_t) \right]_x + \left[ v^d_t \right]_x,
\end{align*}
where
$r^e_t = \log(\exp(r_t) \exp(-r^d_t))$ denotes the rotation error (acting as an element of $\R$ instead of $\so(2)$),
and
$[\cdot]_x, [\cdot]_y$ extract the first and second element of a vector respectively.
Note that $[v^d_t]_y = 0$ for all desired trajectories.
We deploy \GAPS\ to tune the gains $K_1, K_2, K_3, K_4, K_p \in \R_{> 0}^5$
using a logarithmic reparameterization analogous to the quadrotor policy class (\S \ref{sec:log-coords}).

\subsubsection{Cost function}
The cost function
penalizes squared position tracking error
with regularization terms on angular velocity and steering:
\[
	c_t =
	\norm{p_t - p^d_t}_2^2
	+ \recip{30} (\omega_t - \omega^d_t)^2 \\
	+ \recip{15} \tq_t^2.
\]
The regularization weights were chosen empirically to be as small as possible while suppressing oscillations.

\end{document}